\documentclass[conference]{IEEEtran}
\IEEEoverridecommandlockouts
\usepackage{cite}
\usepackage{amsmath,amssymb,amsfonts}
\usepackage{algorithmic}
\usepackage{graphicx}
\usepackage{textcomp}
\usepackage{xcolor}
\usepackage{algorithm}
\usepackage{array}
\usepackage{stfloats}
\usepackage{url}
\usepackage{verbatim}
\usepackage{latexsym}
\usepackage{bm}
\def\BibTeX{{\rm B\kern-.05em{\sc i\kern-.025em b}\kern-.08em
    T\kern-.1667em\lower.7ex\hbox{E}\kern-.125emX}}
		
\begin{document}

\title{Using Autoencoders and AutoDiff to Reconstruct\\Missing Variables in a Set of Time Series\\
\thanks{}
}

\author{\IEEEauthorblockN{1\textsuperscript{st} Jan-Philipp Roche}
\IEEEauthorblockA{\textit{Research, R$\&$D Electronics} \\
\textit{KEB Automation KG}\\
Barntrup, Germany \\
jan-philipp.roche@keb.de}
\and
\IEEEauthorblockN{2\textsuperscript{nd} Oliver Niggemann}
\IEEEauthorblockA{\textit{Professorship of Computer Science} \\
\textit{in Mechanical Engineering} \\
\textit{Helmut-Schmidt-University Hamburg}\\
Hamburg, Germany \\
oliver.niggemann@hsu-hh.de}
\and
\IEEEauthorblockN{3\textsuperscript{rd} Jens Friebe}
\IEEEauthorblockA{\textit{Professorship in Power Electronics} \\
\textit{University of Kassel}\\
Kassel, Germany \\
friebe@uni-kassel.de}
}

\maketitle

\begin{abstract}
Existing black box modeling approaches in machine learning suffer from a fixed input and output feature combination. In this paper, a new approach to reconstruct missing variables in a set of time series is presented. An autoencoder is trained as usual with every feature on both sides and the neural network parameters are fixed after this training. Then, the searched variables are defined as missing variables at the autoencoder input and optimized via automatic differentiation. This optimization is performed with respect to the available features loss calculation. With this method, different input and output feature combinations of the trained model can be realized by defining the searched variables as missing variables and reconstructing them. The combination can be changed without training the autoencoder again. The approach is evaluated on the base of a strongly nonlinear electrical component. It is working well for one of four variables missing and generally even for multiple missing variables.
\end{abstract}


\section{Introduction}\label{intro}
\IEEEPARstart{T}{here} are several black box modeling approaches of time series from nonlinear devices. Such nonlinearities occur, for example, in the voltage and current dependent parameters of many electrical components in electrical circuits. The existing approaches suffer from the fixed input and output feature combination. Important applications such as circuit simulations require a variable combination of input and output features. For the very simple example of a resistor, sometimes the voltage is given and the current is searched or the current is given and the voltage is searched. Transferred to more complex electrical components, there can be a huge number of input and output combinations. Other applications can be equation based hybrid models including neural networks that have to be rearranged and solved for another variable.

In literature, recurrent neural networks (RNN) are used, for example, for the modeling of a CMOS-Receiver, an integrated power amplifier and an integrated CMOS-Logic \cite{else}, for a waveguides and a microstrip low-pass filter \cite{sharma} and for high-speed channels \cite{nuen}. Long short-term memory (LSTM) networks are applied for power amplifiers \cite{gan_amp_lstm} and for nonlinear filters in time domain \cite{iecon}. But neural networks are limited to the combination they are trained to. So a huge number of networks would have to be trained to cover all these cases, which seems neither to be efficient, nor intelligent, nor scalable.

Invertible neural networks (INN) only partially fulfill the required properties. They are typically used to solve inverse problems \cite{INN} \cite{INN_JP} \cite{INN_else} like, for example, reverse measurement-to-parameter-determination. Due to their invertibility, they provide a bijective mapping between the input and output features \cite{INN}. So the inverse combination of features is available. But there is not the full flexibility to vary the combinations of input and output features of the trained model as required for circuit simulations.

Autoencoders \cite{MIT} with every feature on both sides would be suitable to model variable combinations of input and output features if (especially) features on the left side of the trained autoencoder could be eliminated. Unfortunately, eliminating features would lead to a different neural network structure. 

A new approach using autoencoder and automatic differentiation (autodiff) is introduced to overcome these issues. The research questions (RQ) of this work are as follows. Is the presented approach structurally suitable to be a flexible model with variable missing features (RQ1)? It is conceivable that the results differ depending on the actual flexible model configuration. Does the quality of the results depend on the feature that is missing (RQ2)? The approach uses an autoencoder structure and an optimization procedure with autodiff to reconstruct the missing variable. Are the results from the optimized autodiff variable itself better than from the corresponding autoencoder output (RQ3)?

\begin{figure}
\centerline{\includegraphics[width=84mm]{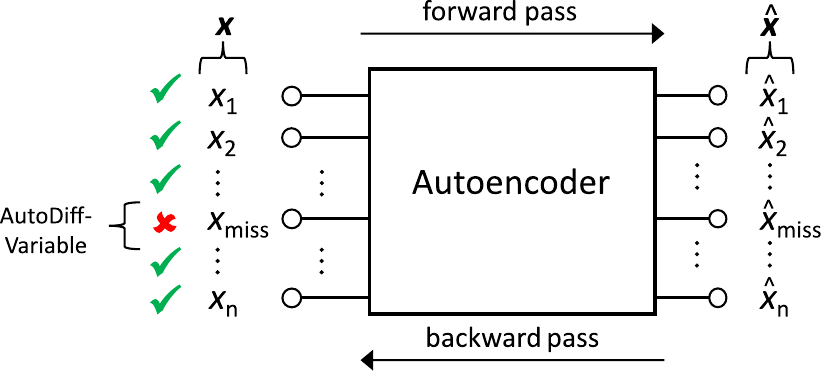}}
\caption{Missing feature set as autodiff variable in autoencoder} \label{introfig}
\end{figure}

Our idea is to define the desired modeling outputs as missing variables. Such a variable corresponds to the output feature of a conventional neural network with fixed features. The missing variable is reconstructed by an optimization process. First, the autoencoder is trained as usual based on complete datasets including every feature. After this training, the neural network parameters of the autoencoder are fixed. Then, the missing feature is set as an autodiff variable (see Fig. \ref{introfig}) so that it can be optimized until the defined loss function reaches a certain minimum. The autodiff variable is optimized based on the unknown application dataset. The utilized loss is calculated by the comparison of the known input feature time series (original data) and the respective autoencoder output. So the variable of the missing feature is optimized until the output of the autoencoder fits the known original data as good as defined by the target loss. It is assumed to find (at least) a local minimum by that procedure and to reconstruct the time series of the missing feature. The concerning time series can be interpreted as the output or result of the modeling procedure. The latent layer of the autoencoder does not need to be considered. To the author's best knowledge, this scalable approach is presented the first time in literature.

It is not the idea to deal with partly missing segments in a time series like, for example, in \cite{miss1} \cite{miss2}. Or to predict an existing time series into the future like mentioned in \cite{nitish}. But completely missing input features are set as autodiff variables. So the known features on left side of the autoencoder are like the inputs, and the autodiff variables on the same side (or the respective ones on the right side) are like the outputs of a conventional model.

The contribution of this paper is as follows: We present a new approach to reconstruct missing variables in a set of time series. The feature of the variable can be changed without training the autoencoder again. Only this variable has to be optimized via autodiff based on the application dataset. We examine the results in dependence of the missing feature and compare them to the fully equipped and trained autoencoder based on an application example. We make an exemplary investigation if the approach can be extended to more than one missing feature.

This paper is structured as follows: In section \ref{rewo}, the related work is outlined. Following, the solution approach including formalization and explanation of the new solution idea is presented. Then, the experiments are described. Finally, the results are discussed, a conclusion is made and an outlook to future work is given.

\section{Related Work}\label{rewo}
Invertible neural networks can change their input and output features. The underlying bijective mapping between the input and output features enables this invertibility \cite{INN}. So INN are typically used to solve inverse problems \cite{INN} \cite{SciPo} \cite{INN_else} like reverse measurement-to-parameter-determination. But the variation of input and output features in time domain simulation of electrical components is not exactly an inverse problem. It requires much more flexibility concerning the configuration of given and searched features.

For the imputation of partly missing segments in time series \cite{sensor_miss}, approaches like mentioned in \cite{miss1} can be applied. These approaches can be statistical models like mean, median or mode imputation. Or machine learning models like linear regression, support vector regression or gradient boosting. From the field of deep learning models, linear memory vector (LIME) gated recurrent neural networks, LIME-LSTM or LIME gated recurrent units are possible approaches. But the imputation of partly missing segments in a time series is a different task compared to the missing of an entire variable in a set of time series.

A LSTM network is used to predict for example stock market prices \cite{lstmstock} \cite{lstmstock2}, which is a future prediction of time series and not exactly meeting our requirements. A LSTM Autoencoder model is applied to reconstruct a video input sequence or to predict the future sequence \cite{nitish}. But such applications reconstruct or predict the same feature, which differs from the required flexibility of input and output combinations.

In \cite{miss2}, convolutional autoencoders are used for waveform imputation of missing data. The trained autoencoder structure estimates the missing segment by averaging from the preceding to the following hidden states. As in the other mentioned approaches, it is about missing data of the same feature and does not address completely missing other features. In our work, completely missing features have to be reconstructed to enable a flexible model with a variable configuration of input and output features. In \cite{miller2}, a state is reconstructed from a variable combination of input sensor channels by using a single flexible convolutional-deconvolutional neural network architecture. The target variable (arterial hypotension) is fixed, a variable combination of input and output features is not addressed.

Deep generative models (DGM) represent probability distributions over multiple variables \cite{MIT}. The three most popular approaches are normalizing flows (NF), variational autoencoders (VAE), and generative adversarial networks (GAN) \cite{introDGM} \cite{impu_zhang} \cite{impu_guo}. They have got the ability to generate new data. For example variational autoencoders can generate new data from the latent space after the distribution is trained \cite{kingma2019introduction}. The generation of new data in itself is not an advantage for reconstructing missing variables in a time series dataset. On the one hand, the generated time series has to meet the real physical behavior in context of the available features in a certain way. On the other hand, deep generative models are instead widely used to impute missing values in (multivariate) time series \cite{e2gan} \cite{impu_zhang} \cite{imputegan} \cite{fang} or to generalize from small labeled to large unlabeled data sets \cite{kingma} and improve semi-supervised learning \cite{maaloe}.

So an approach to build a flexible model with variable missing features to enable a variable input and output configuration is not existing. Furthermore, it is not examined if the quality of the output results depends on the feature that is missing.


\section{Solution Approach}\label{soap}
The solution approach is described in this section. First, the underlying basics are formalized in subsection \ref{forma}. Then, the new solution idea is presented in \ref{nesoid}. It is described in detail in subsection \ref{reco}.

\subsection{Formalization}\label{forma}
An autoencoder consists of an encoder and a decoder. The encoder maps the input $\bm{x}$ to the internal representation $\bm{z}$ in the latent space by the function $f$. The decoder maps $\bm{z}$ to the output $\bm{\hat{x}}$ by the function $g$ \cite{MIT}. Let $\bm{x},\bm{\hat{x}} \in \bm{X} \subseteq \mathbb{R}^n, n \geq 2$. The input is mapped to the output:

\begin{equation}
\bm{\hat{x}}=g(f(\bm{x}))
\label{xtoxhat}
\end{equation}

The functions $f$ and $g$ are represented by neural networks with the parameters $\bm{\theta}$ (weights $\bm{w}$ and biases $\bm{b}$). In the learning process, the following loss function is minimized \cite{MIT}:

\begin{equation}
L(\bm{x}, g(f(\bm{x})))
\label{lossxxhat}
\end{equation}

$L$ is a loss function like for example mean squared error (MSE) which penalizes the difference from $\bm{\hat{x}}$ to $\bm{x}$. The internal representation $\bm{z}$ is not considered any further. 

Automatic differentiation is used to calculate the partial derivative of a function in a certain point. For that, the chain rule is applied in combination with elementary arithmetic operations and functions with regard to the initial parameters \cite{pearlmutter}. The reverse mode of autodiff is preferred here over the forward mode, because the reverse mode is much more efficient (depending on the number of input and outputs \cite{pearlmutter}). Only two passes are required to calculate the function value and the partial derivative through the computation graph \cite{pearlmutter}. The first pass is the forward pass, which calculates the function value (finally $\bm{\hat{x}}$ like in (\ref{xtoxhat})). The second pass is the backward pass. It calculates all the required partial derivatives which is also called the gradient (finally like in (\ref{dLdwb})). 

The forward pass is a calculation with several steps depending on the structure of the neural networks that represent the functions $f$ and $g$ of our autoencoder. The neural network parameters $\bm{\theta}$ are initialized, the input $\bm{x}$ is given here. The output $\bm{\hat{x}}$ is calculated by (\ref{xtoxhat}). 

The backward pass calculates the numerical values of the partial derivatives from the loss function $L$ to the network parameters $\bm{\theta}$. This is processed step by step under the usage of the chain rule. The respective derivative is: 

\begin{equation}
\frac{\partial \: L(\bm{x}, g(f(\bm{x})))}{\partial \:  \bm{\theta}},   \textrm{\:\:for values of } \bm{x}, \bm{\theta}(\bm{w}, \bm{b})
\label{dLdwb}
\end{equation}

\begin{figure}
\centerline{\includegraphics[width=60mm]{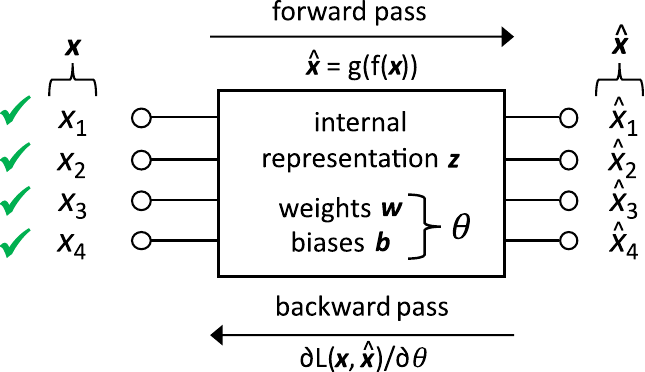}}
\caption{Example of an autoencoder with four features} \label{AEfour}
\end{figure}

An example of an autoencoder with four features and the forward and backward pass is shown in Fig. \ref{AEfour}. Now, the neural network parameters $\bm{\theta}$ can be optimized via optimization methods to minimize the loss function $L$ so that $\bm{\hat{x}}$ becomes more equal to $\bm{x}$. This is used here only to determine the neural network parameters $\bm{\theta}$ for the later reconstruction approach of missing variables.  

In itself, this may seem useless, but the benefit of an autoencoder usually lies in the latent space, which codes the input. The input $\bm{x}$ can be represented by $\bm{z}$ in a simplified way or important features can be extracted \cite{feat_ex1} \cite{feat_ex2}. The output $\bm{\hat{x}}$ is needed to enable the indirect training of the representation $\bm{z}$.

\subsection{New Solution Idea}\label{nesoid}
First, the autoencoder is trained as usual with fully equipped sets of features such as in Fig. \ref{AEfour}. After this training, the neural network of the autoencoder is able to reconstruct the input $\bm{x}$ to the output $\bm{\hat{x}}$ in a certain quality. Now, a missing variable $x_\mathrm{miss}$ in $\bm{x}$ = ($x_\mathrm{1}$, $x_\mathrm{2}$, ..., $x_\mathrm{miss}$, ..., $x_\mathrm{n}$) of an unknown application dataset has to be determined. For that, the network parameters $\bm{\theta}$ are kept constant. The missing variable $x_\mathrm{miss}$ is set as autodiff variable at the input, so that it can be optimized until the loss of all available features reaches a defined minimum. The idea is that $x_\mathrm{miss}$ has to become similar to its original to reach a minimum for the losses of the available features. So the determination of a missing variable is transformed into an optimization task. The approach is explained for multiple missing variables and more detailed in subsection \ref{reco}.

\subsection{Reconstruction of Missing Variables}\label{reco}
The considerations from subsection \ref{forma} have to be modified to implement the new solution idea from subsection \ref{nesoid}. The missing variables are called $\bm{x}_\mathrm{miss}$ and set as autodiff variables. The forward pass from (\ref{xtoxhat}) is additionally described by:

\begin{equation}
\bm{\hat{x}}=g(f(\bm{x})), \:\:\:\: \bm{x}_\mathrm{miss} \in \bm{x}
\label{xtoxhatxmiss}
\end{equation}

For the backward pass, only the available features can be used to calculate the loss. The input is reduces by $\bm{x}_\mathrm{miss}$ to:

\begin{equation}
\bm{x}_\mathrm{red}=\bm{x} \setminus \bm{x}_\mathrm{miss}
\label{xred}
\end{equation}

And the output analogous to:

\begin{equation}
\bm{\hat{x}}_\mathrm{red}=\bm{\hat{x}} \setminus \hat{\bm{x}}_\mathrm{miss}
\label{xhatred}
\end{equation}

It is of central importance that the backward pass is done with relation to the missing variables $\bm{x}_\mathrm{miss}$. 
And also taking (\ref{xred}) and (\ref{xhatred}) into account, (\ref{dLdwb}) changes to:

\begin{equation}
\frac{\partial \: L_\mathrm{red}(\bm{x}_\mathrm{red}, \bm{\hat{x}}_\mathrm{red})}{\partial \: \bm{x}_\mathrm{miss}}
\label{dLavdwbav}
\end{equation}

The loss is defined here with reference to \ref{xred} and \ref{xhatred}. So the partial derivative in (\ref{dLavdwbav}) can now be used to optimize the missing variables $\bm{x}_\mathrm{miss}$ in context of the previously trained and fixed autoencoder. The optimization task can be formulated as:

\begin{equation}
\min_{{\bm{x}_\mathrm{miss}}} L_\mathrm{red}(\bm{x}_\mathrm{red}, \bm{\hat{x}}_\mathrm{red}), \:\:\:\: \bm{x}_\mathrm{miss} \in \bm{x}
\label{mintask}
\end{equation}

\begin{figure}
\centerline{\includegraphics[width=86mm]{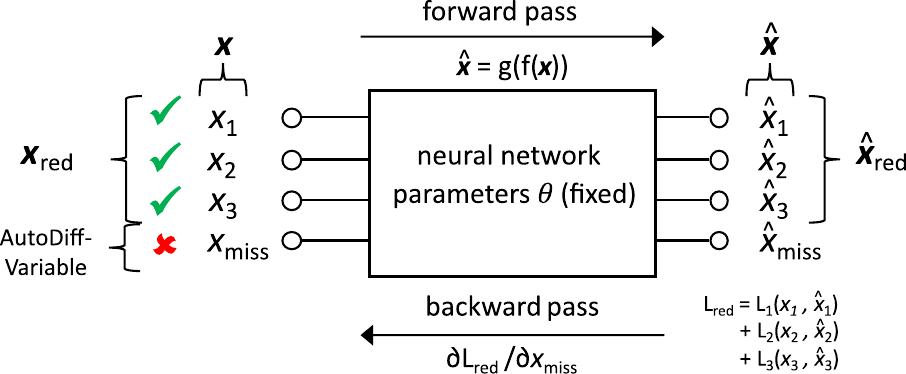}}
\caption{Modified autoencoder usage with one of four features missing(autodiff variable $x_\mathrm{miss}$)} \label{AEnew}
\end{figure}

This optimization process is finished when the outputs for the available features fit the concerning original data in a defined grade (loss). The modified autoencoder usage is shown in Fig. \ref{AEnew} by example of one of four features missing.

The respective loss is defined as:

\begin{equation}
L_\mathrm{red}(\bm{x}_\mathrm{red}, \bm{\hat{x}}_\mathrm{red}) = L_{1}(\bm{x}_{1}, \bm{\hat{x}}_{1}) + L_{2}(\bm{x}_{2}, \bm{\hat{x}}_{2}) + L_{3}(\bm{x}_{3}, \bm{\hat{x}}_{3})
\label{Lavdef}
\end{equation}

If $x_\mathrm{miss}$ would be perfectly equal to the original data, the losses for all features would be equal to the ones of the finished training process. So the finished training process with fully known features is the reference for the reconstruction process by (\ref{dLavdwbav}). It is assumed that this result will not be perfectly met. But a sufficiently good local minimum is aimed. The neural network parameters $\bm{\theta}$ are kept constantly as trained before. The representation $\bm{z}$ in the latent space is not actively used here. An extension to multiple missing variables as introduced in this subsection is generally possible. Autodiff is not limited to a single optimization variable, so the approach is easily scalable. Furthermore, autodiff is freely available in tools like PyTorch, which enables a widely usage.

\section{Experiments}\label{expe}
\begin{figure}[t]
\centerline{\includegraphics[width=70mm]{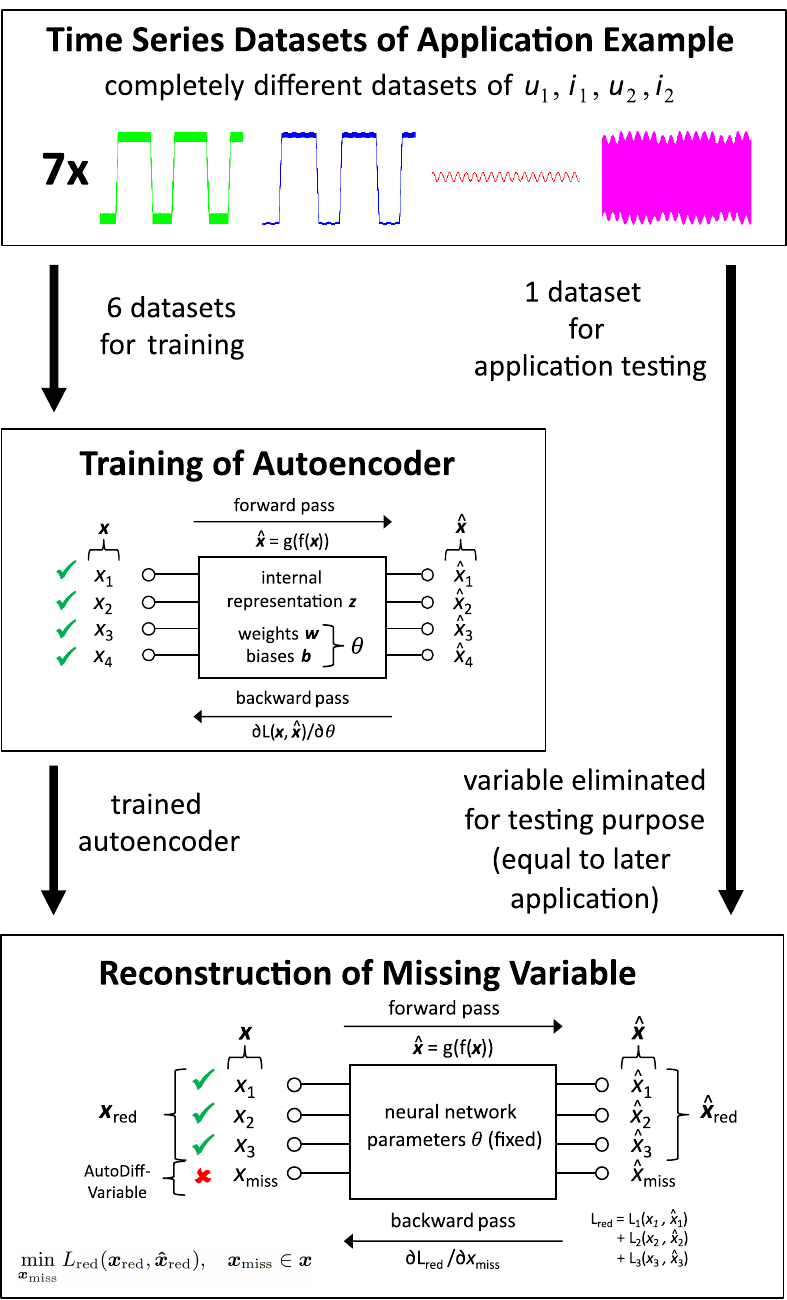}}
\caption{Overall procedure of reconstructing one missing variable using autoencoder and autodiff by hand of application example. Six datasets of currents and voltages are used for the conventional training of the autoencoder. The trained autoencoder parameters $\bm{\theta}$ are fixed and the separate testing dataset used to test the reconstruction of the missing variable $x_\mathrm{miss}$. This missing variable was previously eliminated from the dataset for testing purpose. The later application of the reconstruction method is simulated by that.} \label{over}
\end{figure}

The performed experiments are described in this section. First, the used neural network structure is explained. Then, the used application example is introduced. After that, the training of the autoencoder is presented including the hyperparameters. And finally, the solution for reconstructing the missing input feature is described. The overall procedure is shown in Fig. \ref{over}.

\subsection{Neural Network Structure}
\begin{figure}
\centerline{\includegraphics[width=86mm]{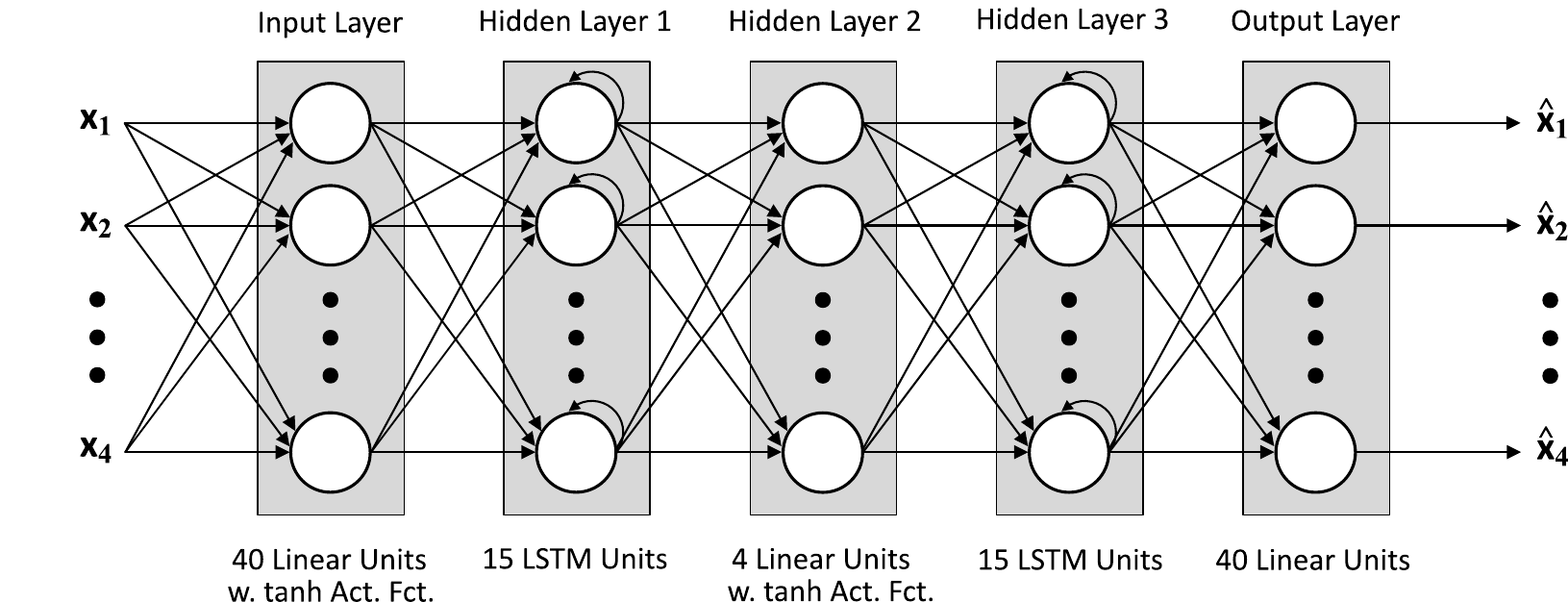}}
\caption{Neural network structure of the used autoencoder (not unfolded representation)} \label{NN}
\end{figure}

The used neural network structure has got the shape of an hourglass, which comes from the usual application of autoencoders. It has got an input layer, three hidden layers (including the latent space in the middle) and an output layer. The first and the third hidden layers are LSTM layers, because they are suitable to represent time dependent behaviour of sequential data. They contain a memory and forget functionality, so that a trained short-term memory lasts a long time \cite{21iecon}. Such kind of behavior is suitable to model the time dependent behavior of electrical circuits like, for example, nonlinear filters \cite{iecon}. The other layers are linear layers with a tanh activation function, except for the output layer. The neural network structure is shown in Fig. \ref{NN}. In detail, it is defined by empirical-iterative experiments.

A sliding window is applied to all features for training the time dependency. The features time series are transformed into a tensor while a sliding window with the width of 3 time steps is applied. So the sequence length is 3. More details can be found in \cite{iecon}. The hyperparameters and datasets are described in subsection \ref{pret} and \ref{remi}.

\subsection{Application Example}
\begin{figure}
\centerline{\includegraphics[width=68mm]{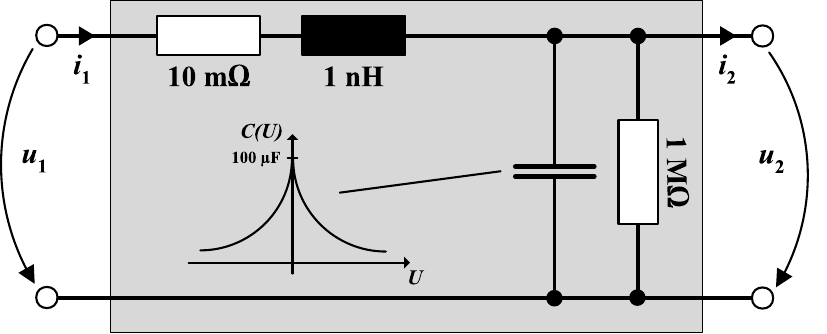}}
\caption{Application example (strongly nonlinear filter circuit); different combinations of available and missing features are possible (\textit{u}\textsubscript{1}, \textit{u}\textsubscript{2}, \textit{i}\textsubscript{1}, \textit{i}\textsubscript{2})} \label{AE1}
\end{figure}

A nonlinear filter circuit is used as application example. It contains a capacitance that is strongly dependent on the respective voltage and has been used before in \cite{iecon} and \cite{epe}. The application example including the related voltages and currents is shown in Fig. \ref{AE1}. In a typical application (circuit simulation), the given and missing features are not fixed.

This nonlinear filter is used here, because it is strongly nonlinear, but still a simple structure. It is not in the application focus of the new approach, but it provides a handable example to introduce the approach. The voltages \textit{u}\textsubscript{1} and \textit{u}\textsubscript{2} and the currents \textit{i}\textsubscript{1} and \textit{i}\textsubscript{2} used as features in the following (sub)sections. To generate fully equipped datasets, the nonlinear filter circuit is simulated in LTspice \cite{ltspice}. Multiple typical time series characteristics are applied to the filter to generate different datasets over a wide range of working points. Such characteristics of these voltages and currents are, for example, combinations of DC, low frequency sine, high frequency sine and trapezoidal waveforms. After the simulation in LTspice, the time series of the four features are made equidistant with a sample rate of 100 Mpts/s.

\subsection{Training} \label{pret}
For the training, the autoencoder from Fig. \ref{AEfour} is trained with six fully equipped datasets. The four features of $\bm{x}$ are used for the voltages \textit{u}\textsubscript{1} and \textit{u}\textsubscript{2} and the currents \textit{i}\textsubscript{1} and \textit{i}\textsubscript{2} of the nonlinear filter example. The training is performed for 1000 epochs. A target loss was not defined here. The training is implemented in PyTorch \cite{pytorch} and run on a graphics processing unit (NVIDIA GeForce RTX 3090). Every epoch for every dataset takes about 17s runtime.
All batches of one time series dataset are fed to the autoencoder before the update of the parameters $\bm{\theta}$ is executed. Furthermore, the six time series are always alternated again to prevent from the domination of the last trained dataset. This procedure yielded the best results in similar training tasks \cite{iecon}. 
The data is rescaled with MinMaxScaler, Adam is used as optimizer, MSE is the chosen loss function and the default learning rate of 0.001 worked reliably. Variations did not lead to any notable advantage.

\subsection{Reconstruction of Missing Input Features} \label{remi}
Now, the proposed solution approach from section \ref{soap} is applied to the trained autoencoder. An unknown dataset with a missing feature is fed to the structure as shown in Fig. \ref{AEnew}. The missing variable $x_\mathrm{miss}$ can be \textit{u}\textsubscript{1}, \textit{u}\textsubscript{2}, \textit{i}\textsubscript{1} or \textit{i}\textsubscript{2}. In the experiments, this will be varied to test the approach with regard to the different features. The missing variable $x_\mathrm{miss}$ is initialized with zeros. Experiments with different initialization values did not show any notable advantage. The training is performed as shown in Fig. \ref{AEnew} and described in subsection \ref{reco}. 
The scaler, the optimizer and the loss function are kept the same in comparison to the training (MinMaxScaler, Adam, MSE). The learning rate is increased to 0.005, which showed good results and speeds up the optimization process in the application usage. The optimization variable is changed from $\bm{\theta}$ to $x_\mathrm{miss}$ and the loss calculation is adapted to the available features. The procedure of transforming to the tensors and feeding it to the autoencoder is also kept the same. The only difference is that the missing input variable $x_\mathrm{miss}$ is optimized towards a single dataset, which is the application dataset. 
As in the training, the suitable number of epochs is evaluated by hand and results in 300 epochs (about 21s runtime each). A fixed target loss is also not defined here. The optimization procedure in this subsection has to be repeated if the feature of the missing variable is changed. A procedure run with two missing variables at the same time is also performed, but is not in the primary focus of this work. For this, the hyperparameters are kept the same. Only the number of epochs has to be increased to 3000 (about 21s runtime each).


\section{Results and Discussion}\label{resdi}
\begin{figure}
\centerline{\includegraphics[width=86mm]{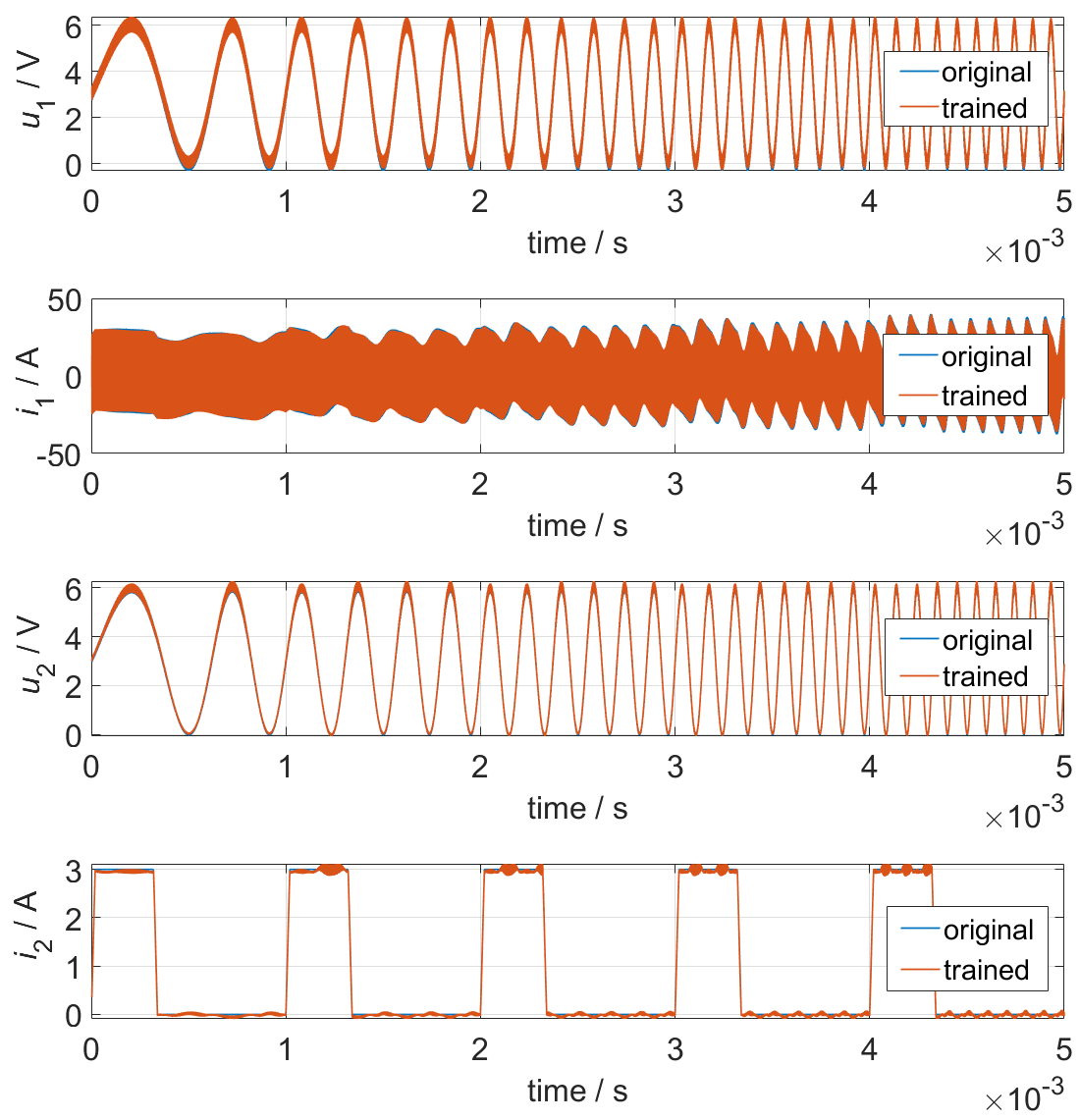}}
\caption{Training results of unknown test dataset compared to original data (see Fig. \ref{AEfour} and Fig. \ref{over})} \label{res_SS_pre}
\end{figure}

The results of the training are shown in Fig. \ref{res_SS_pre}. These training results are the reference for the later evaluation of the reconstruction results for the missing variable. For \textit{u}\textsubscript{1}, \textit{i}\textsubscript{1} and \textit{u}\textsubscript{2}, there is a very good fit in time domain. The training results for \textit{i}\textsubscript{2} fit in principle but show some deviations. These deviations are characteristically similar to the time curves of \textit{i}\textsubscript{1}.

\begin{figure}
\centerline{\includegraphics[width=86mm]{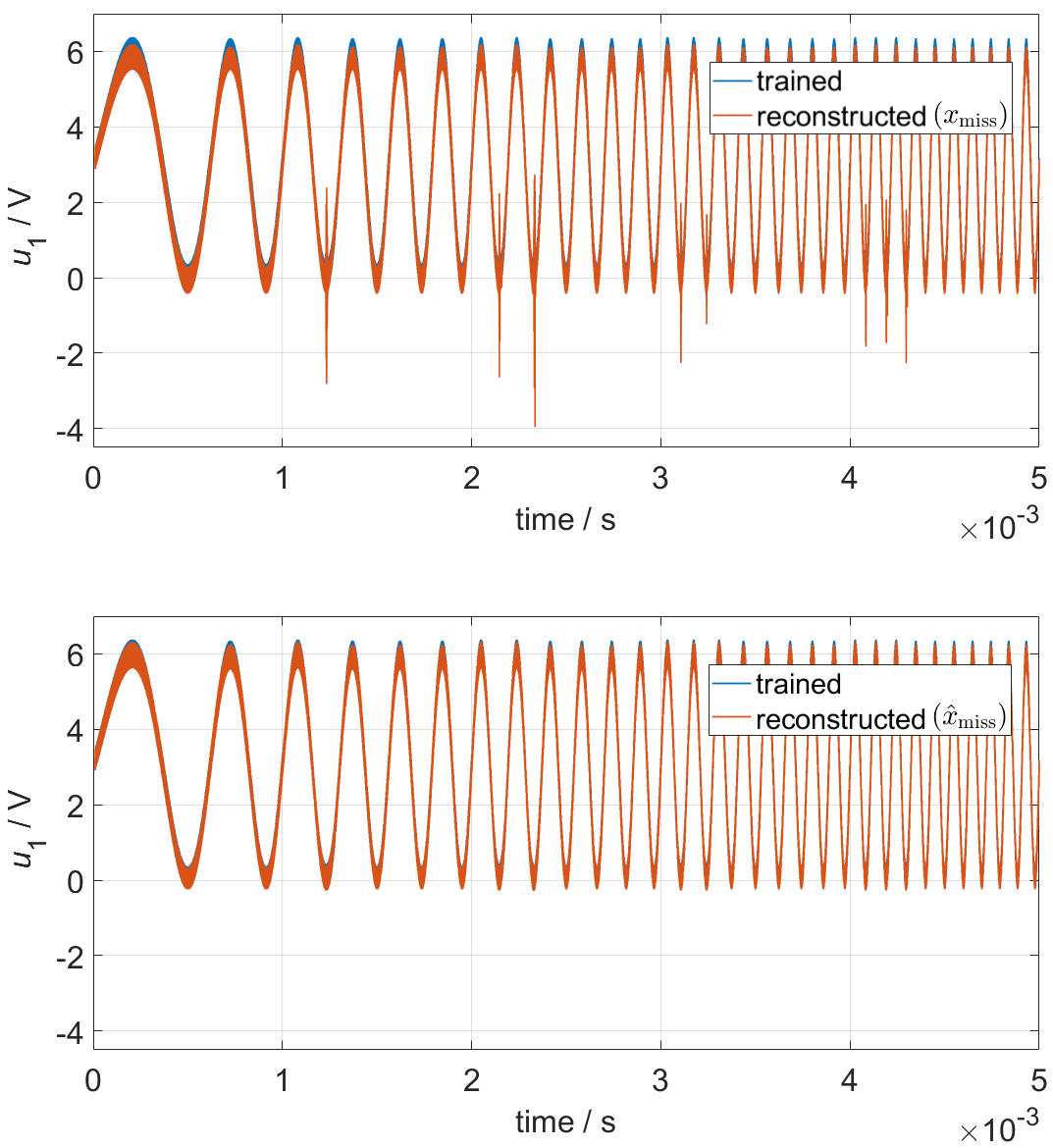}}
\caption{Comparison of $x_\mathrm{miss}$ and $\hat{x}_\mathrm{miss}$ to training data for reconstruction results of \textit{u}\textsubscript{1}} \label{reconVinADVar1}
\end{figure}

The reconstruction of the missing variable is tested with an unknown dataset as shown in Fig. \ref{over} and explained in subsection \ref{remi}. In Fig. \ref{reconVinADVar1}, the reconstruction for \textit{u}\textsubscript{1} is compared to the training reference. An additional forward pass of the reconstructed $x_\mathrm{miss}$ leads to $\hat{x}_\mathrm{miss}$, which shows a better fit. The time curve of $x_\mathrm{miss}$ shows some additional spikes. It is assumed, that the reason for this effect is the loss definition in (\ref{Lavdef}) which refers to $\bm{\hat{x}}_\mathrm{red}$ after the forward pass. In the following, the results $\hat{x}_\mathrm{miss}$ after the forward pass are used.

\begin{figure}
\centerline{\includegraphics[width=86mm]{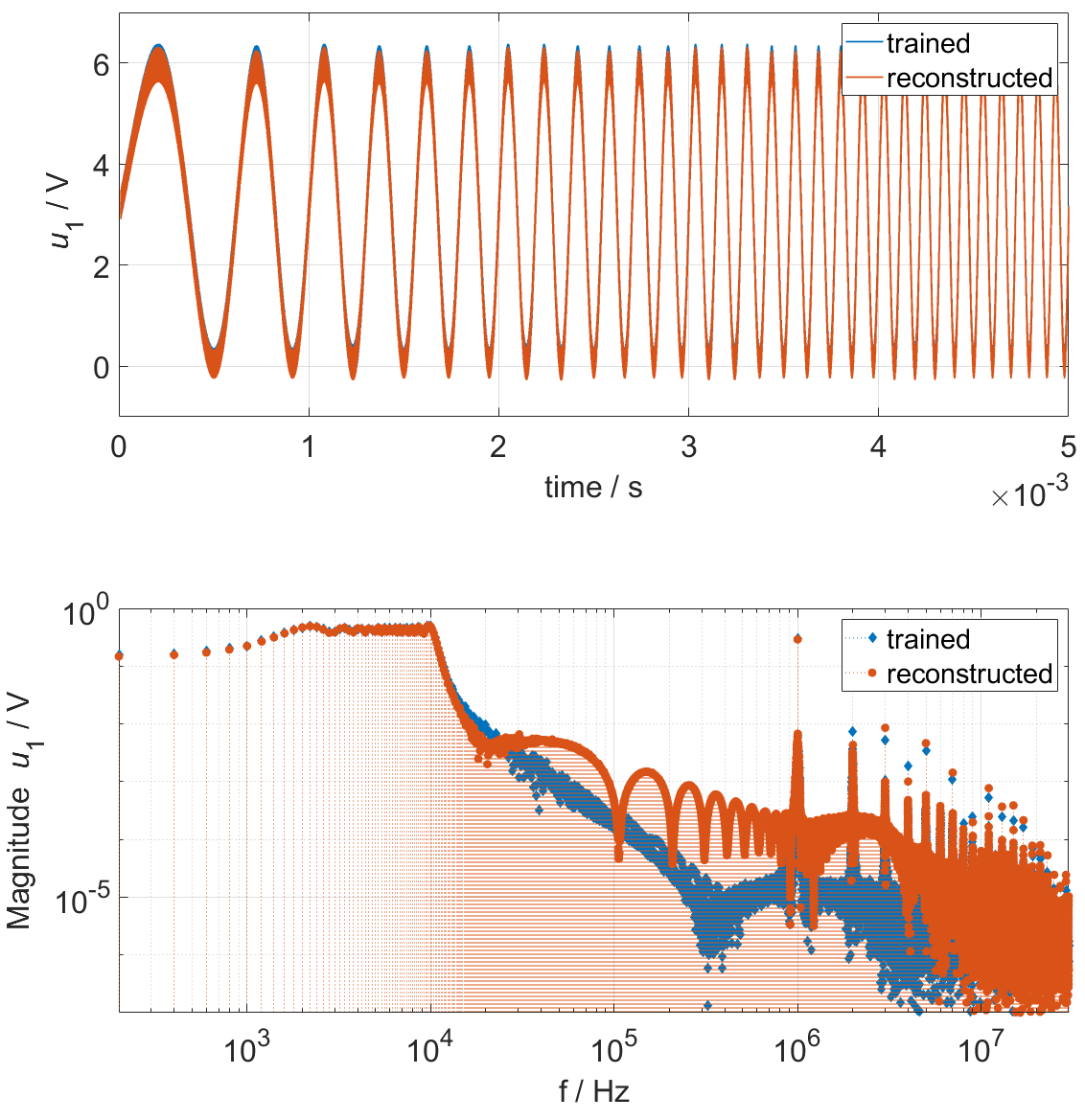}}
\caption{Reconstruction results of \textit{u}\textsubscript{1} compared to training data} \label{reconVin}
\end{figure}

\begin{figure}
\centerline{\includegraphics[width=86mm]{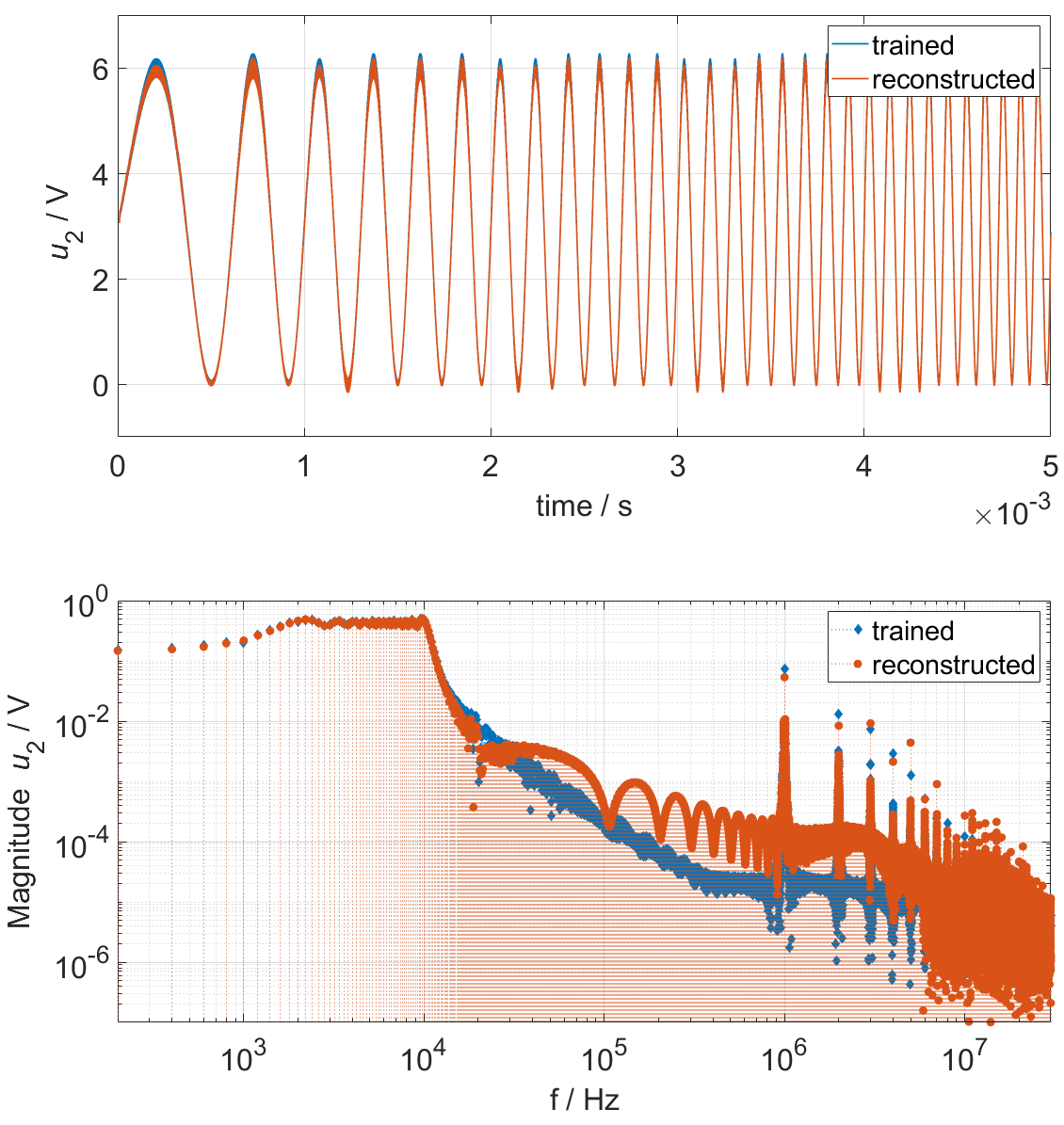}}
\caption{Reconstruction results of \textit{u}\textsubscript{2} compared to training data} \label{reconVout}
\end{figure}

\begin{figure}
\centerline{\includegraphics[width=86mm]{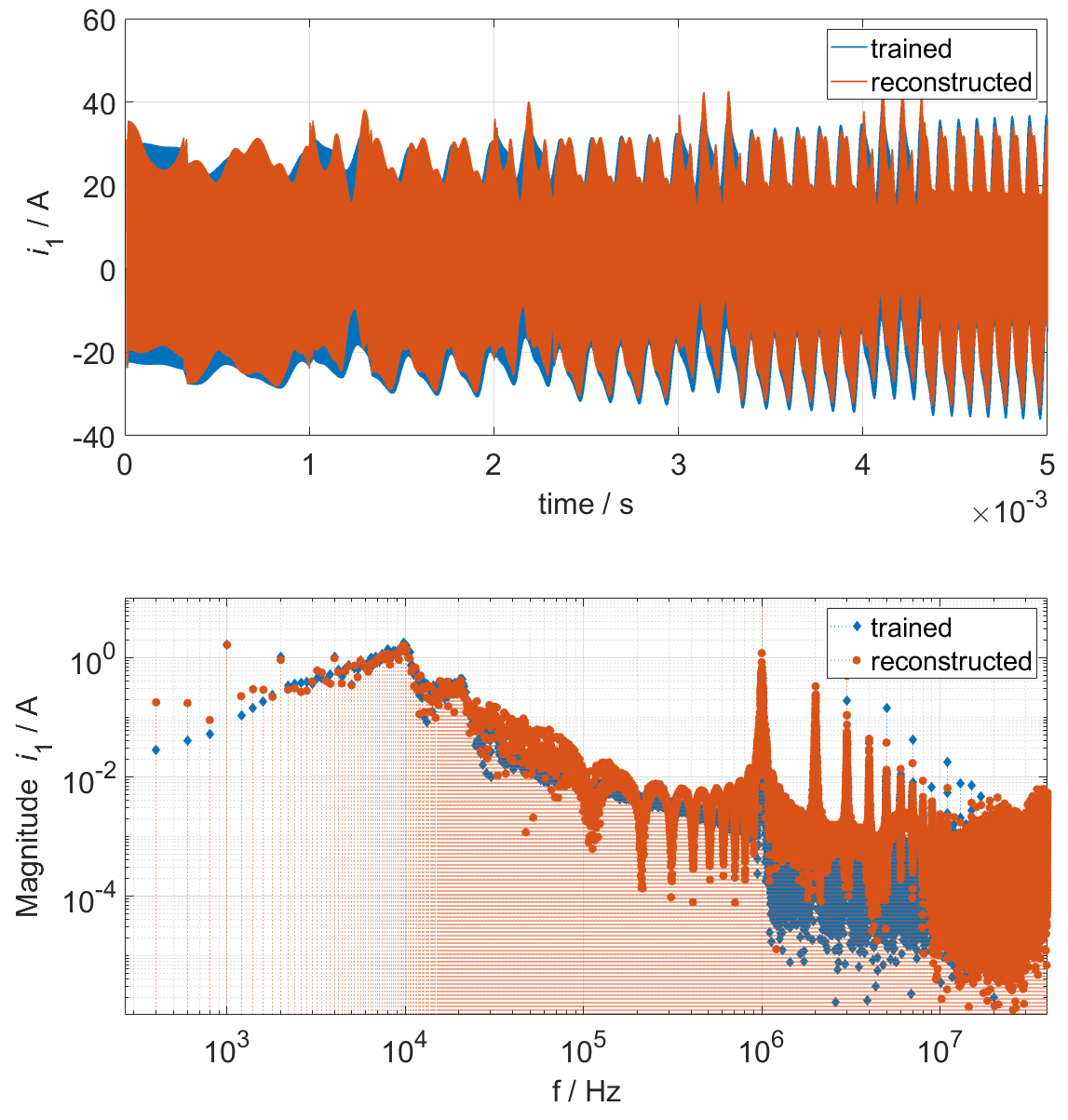}}
\caption{Reconstruction results of \textit{i}\textsubscript{1} compared to training data} \label{reconIin}
\end{figure}

\begin{figure}
\centerline{\includegraphics[width=86mm]{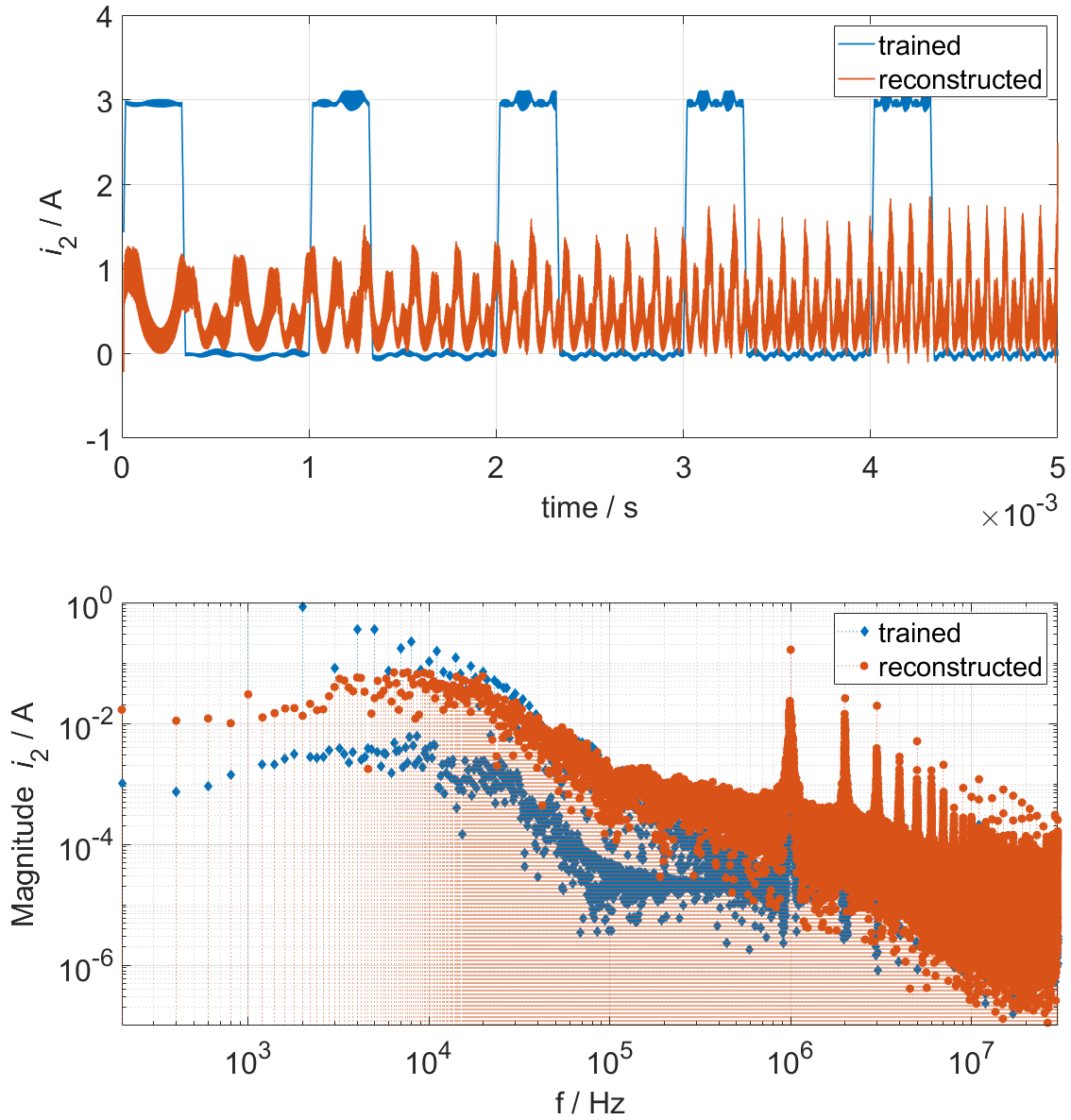}}
\caption{Reconstruction results of \textit{i}\textsubscript{2} compared to training data} \label{reconIout}
\end{figure}

The results for the reconstructed \textit{u}\textsubscript{1}, \textit{i}\textsubscript{1}, \textit{u}\textsubscript{2} and \textit{i}\textsubscript{2} of the unknown test dataset are compared to the training and shown in Fig. \ref{reconVin}, Fig. \ref{reconVout}, Fig. \ref{reconIin} and Fig. \ref{reconIout}. The respectively shown variable was missing and the other ones available. So each of these figures represents the result of an individual reconstruction process (Fig. \ref{AEnew}).

In Fig. \ref{reconVin}, the time domain result for \textit{u}\textsubscript{1} is almost congruent with the reference. In the frequency domain, the amplitudes up to 10 kHz show a very good fit. At the high frequencies, the dominant amplitudes are well represented. In the middle range, there are more deviations. The results for \textit{u}\textsubscript{2} in Fig. \ref{reconVout} are very similar.

The time curve of the reconstructed \textit{i}\textsubscript{1} in Fig. \ref{reconIin} represents the main characteristics in comparison to the reference. There is a congruent fit over the entire frequency range with some deviations concerning the dominant amplitudes in the high frequency range.

For the reconstruction of \textit{i}\textsubscript{2}, the results in Fig. \ref{reconIout} show a completely different time curve. The reconstruction does not fit to the reference at all. The reconstructed time curve shows characteristic parts of \textit{i}\textsubscript{1}. At a lower level, this has been an issue in the training of \textit{i}\textsubscript{2}. So the quality of the training seems to be very important for the reconstruction result of the missing variable. The effect already seen in the training is much more dominant in the reconstruction process. 

\begin{figure}
\centerline{\includegraphics[width=86mm]{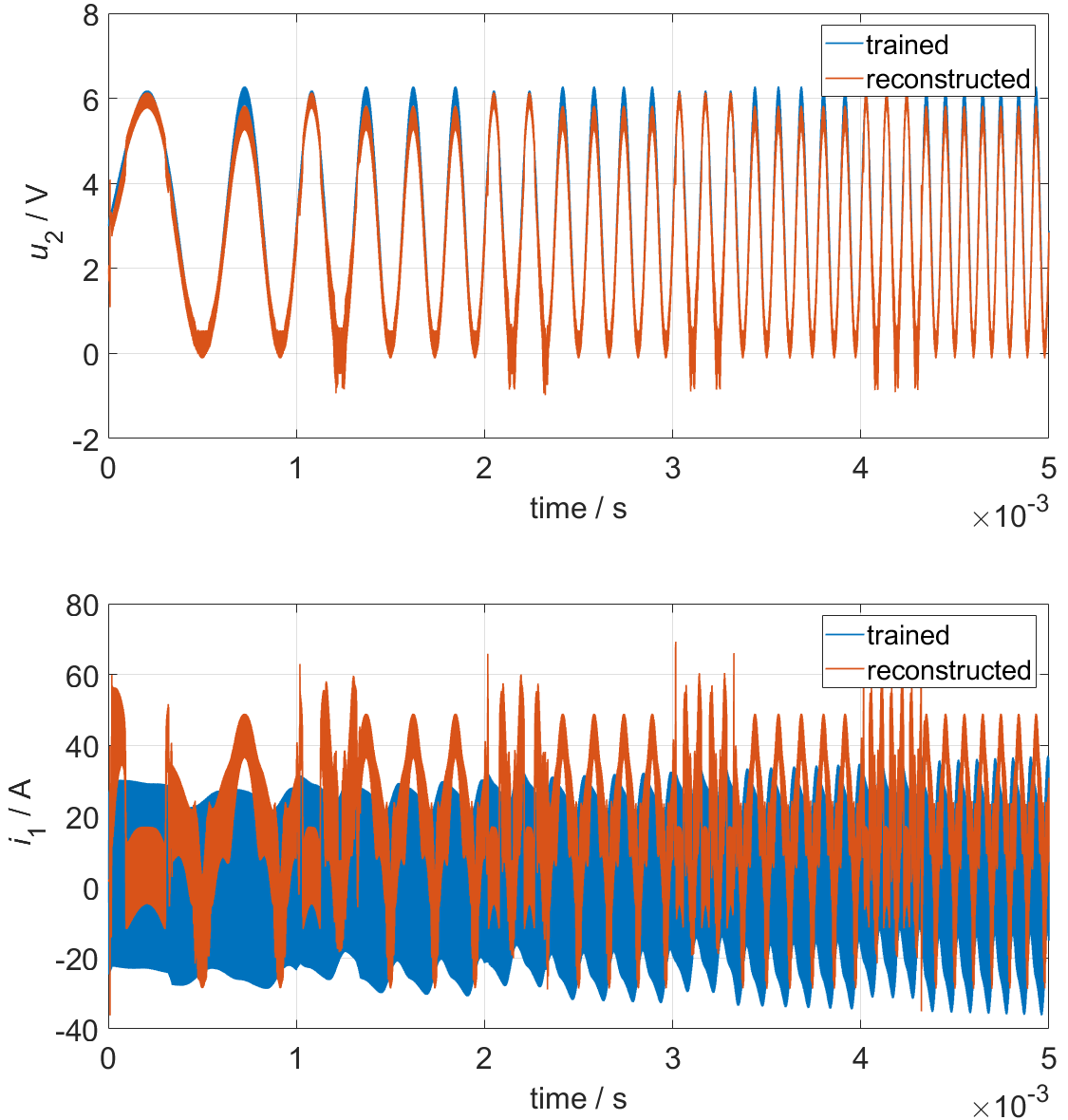}}
\caption{Reconstruction results of \textit{u}\textsubscript{2} and \textit{i}\textsubscript{1} compared to training data (two missing variables reconstructed at the same time)} \label{reconVoutIin}
\end{figure}

In Fig. \ref{reconVoutIin}, an exemplary result for a reconstructions process extended to two missing variables at the same time is shown. The reconstructed time curve of \textit{u}\textsubscript{2} shows a bit more deviations compared to the respective result for the reconstruction of only a single missing variable. But the general characteristic is likewise well represented. In contrast to this, the deviations for \textit{i}\textsubscript{1} are much larger. In general, it can be said that the reconstruction of more than a single missing feature is also working, but the deviations are larger. It has to be kept in mind that the missing variables are initialized with zeros and optimized. So the time curve of \textit{i}\textsubscript{1} is optimized in the right direction but still does not reach a satisfying fitting.

The reconstruction procedure itself is working for a missing variable, even with very little training data. And generally also for multiple missing variables. It has to be figured out how the quality of the results can be improved, for example by the usage of more training data, by hyperparameter tuning or modifications of the autoencoders neural network structure. Furthermore, the quality of the training seems to be especially important for a successful reconstruction process. A correlation between the quality of the training and reconstruction can be found in the presented results. To improve the training process, it is recommended to modify the loss function to focus on the fitting of (in this case) \textit{i}\textsubscript{2}. Even during the optimization of the autodiff variable for reconstruction, it is recommended to put a focus on the variable that is difficult to reconstruct.


\section{Conclusion and Future Work}\label{cofu}
The presented approach is suitable to be a flexible model with variable missing features (RQ1). After the training, the autoencoder is fixed and the missing feature set as autodiff variable and reconstructed by optimization. The quality of the result depends on the reconstructed feature (RQ2). There is a correlation between the quality of the training and the reconstruction for the different features. The result of the reconstructed feature fits better after a forward pass trough the autoencoder (RQ3). It is not recommended to use the optimized autodiff variable itself. 

Thus, the presented new solution idea allows to flexibly combine the input and output features of a model without the need to train the neural network of the autoencoder again. Only the missing feature has to be optimized via autodiff based on the application dataset. Even the reconstruction of two (or potentially more) missing features at the same time is principally working, although some improvements still must to be achieved. 

For future work, it is recommended to tune the hyperparameters and to use more datasets for training. A stopping criterion for the reconstruction process is missing, because a usual validation loss can not be utilized due to the missing feature. Eventually, a kind of prediction process can be implemented to create data of the missing feature for a validation loss calculation. 
Even the structure of the neural network should be optimized in future work. With much more datasets for training, a larger neural network can be used and the results potentially improved. The extension to more missing features has to be evaluated more detailed. 
The latent space of the autoencoder is not actively used here and should be examined. Perhaps, the reconstruction performance can be improved by a better understanding of the latent space function in this approach. The optimal size of the latent space can be dependent on the number of the application features. Or on the number of the systems state space variables which is dependent, for example, on the electrical circuits number of energy storage elements.

\section*{Acknowledgment}

Parts of this work were funded by the Deutsche Forschungsgemeinschaft (DFG, German Research Foundation) – Project-ID 502998206.

Parts of this work are supported bei KEB Automation KG.


\newpage

\begin{IEEEbiography}[{\includegraphics[width=1in,height=1.25in,clip,keepaspectratio]{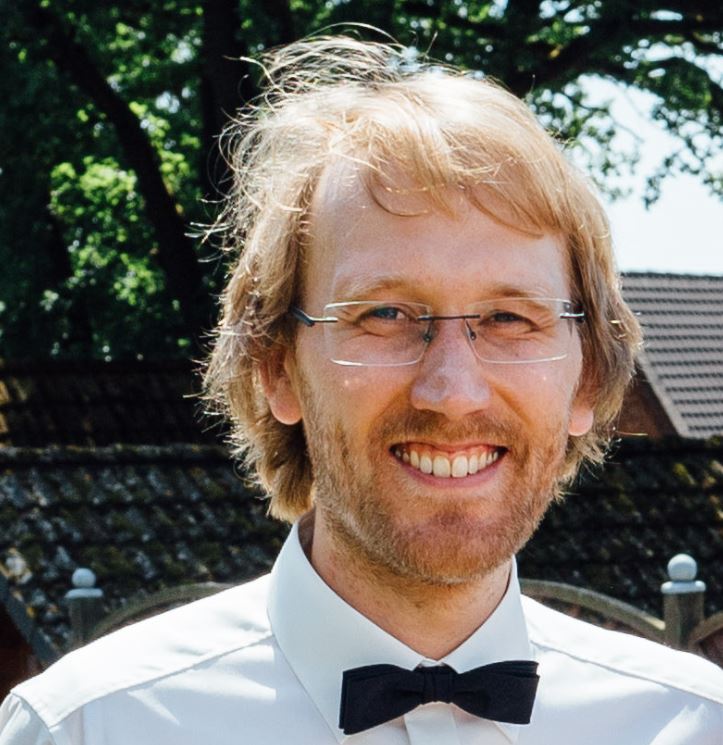}}]{Jan-Philipp Roche} (Graduate Student Member, IEEE) was born in Lemgo, Germany, in 1987. He received the B.Sc. degree in electrical engineering from the Hochschule Ostwestfalen-Lippe (University of Applied Sciences) in 2011, and the M.Sc. degree in electrical engineering from the FernUniversität in Hagen (University) in 2018. The master thesis was done in a corporation between KEB Automation KG, FernUniversität in Hagen and the Institute for Drive Systems and Power Electronics at Leibniz University Hannover. Actually, he is enrolled as a doctoral student at Leibniz University Hannover. He is researcher and former power electronics engineer at KEB Automation KG, where he started as a student trainee in 2008. He worked on industrial inverters in the power range of several hundred kW. Currently he works on the EMC of power electronics.

\end{IEEEbiography}

\vskip 0pt plus -1fil

\begin{IEEEbiography}[{\includegraphics[width=1in,height=1.25in,clip,keepaspectratio]{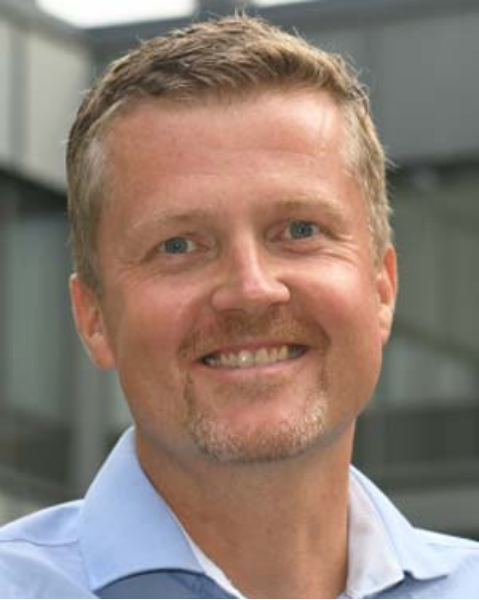}}]{Oliver Niggemann} (Member, IEEE) received the Ph.D. degree in visual data mining of graph-based data from Paderborn University, Paderborn, Germany, in 2001. He, then, worked for eight years in leading positions in industry. From 2008 to 2019, he had a professorship with the Institute for Industrial Information Technologies, Lemgo, Germany. Until 2019, he was also the Deputy Head of the Fraunhofer IOSB-INA, which works in industrial automation. On April 1, 2019, he took over the university professorship of Computer Science in mechanical engineering with Helmut Schmidt University, Hamburg, Germany, where he is conduction research at the Institute for Automation Technology IfA in the field of artificial intelligence and machine learning for cyber-physical systems. 
\end{IEEEbiography}

\vskip 0pt plus -1fil

\begin{IEEEbiography}[{\includegraphics[width=1in,height=1.25in,clip,keepaspectratio]{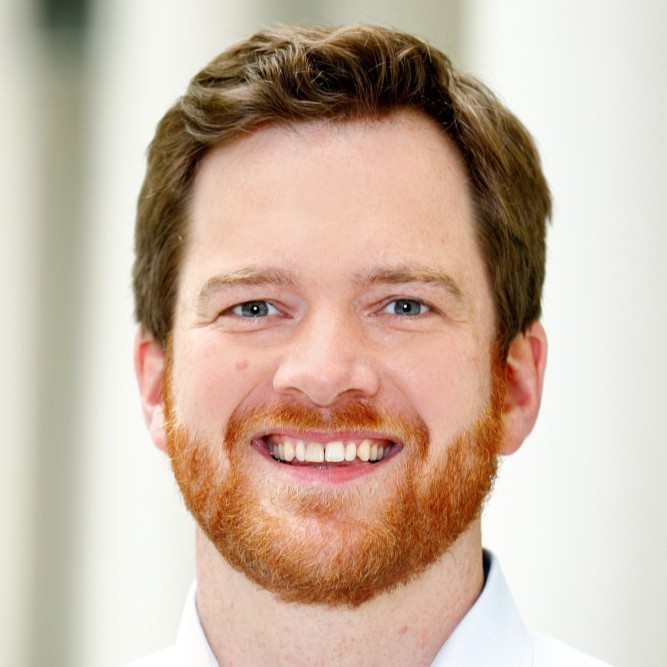}}]{Jens Friebe} (Senior Member, IEEE) was born in Göttingen, Germany. He received the B.Sc., M.Sc., and Dr.-Ing. degrees in electrical engineering from the University of Kassel, Germany. He has been responsible for the research area of passive components in power electronics at the Institute for Drive Systems and Power Electronics, Leibniz University Hannover, Germany, since January 2018. Before that, he worked for more than 13 years at SMA Solar Technology, Germany, in the field of PV-inverter topologies, wide-bandgap semiconductors, magnetic components, control strategies for high switching frequencies, and power electronics packaging. He has invented more than 30 granted patents in the field of power electronics.
\end{IEEEbiography}

\end{document}